\documentclass[letterpaper]{article} 
\usepackage{aaai24}  
\usepackage{times}  
\usepackage{helvet}  
\usepackage{courier}  
\usepackage[hyphens]{url}  
\usepackage{graphicx} 
\usepackage{natbib}  
\usepackage{caption} 
\usepackage{algorithm}

\usepackage{newfloat}
\usepackage{listings}
\DeclareCaptionStyle{ruled}{labelfont=normalfont,labelsep=colon,strut=off} 
\floatstyle{ruled}
\newfloat{listing}{tb}{lst}{}
\floatname{listing}{Listing}
\nocopyright 

\usepackage{amsmath}
\usepackage{amssymb}
\usepackage{booktabs}

\usepackage{amsmath,amsfonts}
\usepackage{pifont} 

\newcommand{\cmark}{\ding{51}}
\newcommand{\xmark}{\ding{55}}

\usepackage{tikz}
\usepackage{comment}

\usepackage{color}
\usepackage{multirow}
\usepackage{caption}

\usepackage{pifont}
\usepackage{mathtools}
\usepackage[noend]{algpseudocode}

\usepackage{subcaption}
\usepackage{tabularx}

\usepackage[capitalize]{cleveref}
\crefname{section}{Sec.}{Secs.}
\Crefname{section}{Section}{Sections}
\Crefname{table}{Table}{Tables}

\newcommand{\etal}{\textit{et al.}}
\newcommand{\eg}{\textit{e.g.}}
\newcommand{\ie}{\textit{i.e.}}

\title{Deep Deformable Models: Learning 3D Shape Abstractions with Part Consistency}
\author{
    Di Liu\textsuperscript{\rm 1},
    Long Zhao\textsuperscript{\rm 2},
    Qilong Zhangli\textsuperscript{\rm 1},
    Yunhe Gao\textsuperscript{\rm 1},
    Ting Liu\textsuperscript{\rm 2},
    Dimitris N. Metaxas\textsuperscript{\rm 1}
}
\affiliations{
    \textsuperscript{\rm 1}Rutgers University,
    \textsuperscript{\rm 2}Google Research

}

\usepackage{bibentry}

\begin{document}

\maketitle

\begin{abstract}
The task of shape abstraction with semantic part consistency is challenging due to the complex geometries of natural objects. Recent methods learn to represent an object shape using a set of simple primitives to fit the target. \textcolor{black}{However, in these methods, the primitives used do not always correspond to real parts or lack geometric flexibility for semantic interpretation.} In this paper, we investigate salient and efficient primitive descriptors for accurate shape abstractions, and propose \textit{Deep Deformable Models (DDMs)}. DDM employs global deformations and diffeomorphic local deformations. These properties enable DDM to abstract complex object shapes with significantly fewer primitives that offer broader geometry coverage and finer details. DDM is also capable of learning part-level semantic correspondences due to the differentiable and invertible properties of our primitive deformation. 
Moreover, DDM learning formulation is based on dynamic and kinematic modeling, which enables joint regularization of each sub-transformation during primitive fitting. 
Extensive experiments on \textit{ShapeNet} demonstrate that DDM outperforms the state-of-the-art in terms of reconstruction and part consistency by a notable margin. 
\end{abstract}

\begin{figure*} [t]
\begin{center}
\includegraphics[width=1\linewidth]{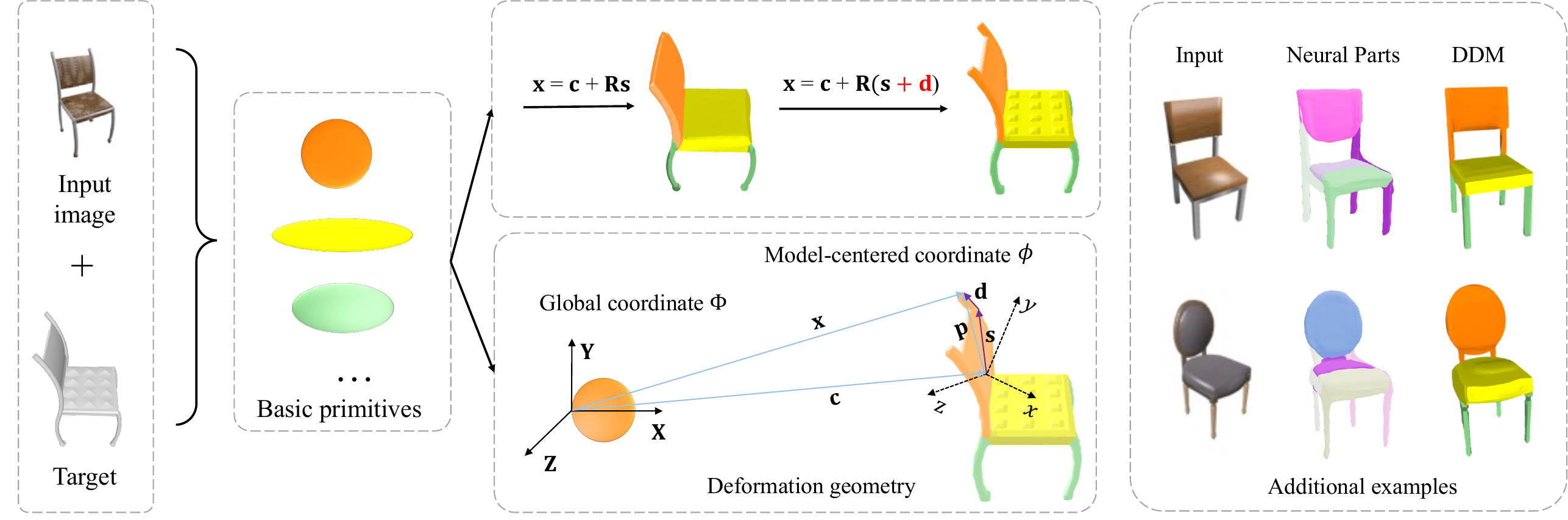}
\end{center}
\caption{\textbf{DDM dynamic fitting and sample reconstruction.} Given a single 2D image and a 3D target shape, we deform a small number of primitives to fit different parts of the object shape via translation $\textbf{c}$, rotation $\textbf{R}$ as well as deformations $\textbf{s} + \textbf{d}$. Our model is able to predict high-fidelity and semantically consistent 3D object shapes with fine details.}
\label{cover_framework}
\end{figure*}
\section{Introduction}
Accurately abstracting complex object shapes with a set of primitives that offer semantic interpretability has been a long-standing goal in computer vision, medical image analysis and graphics. It can be used in a variety of downstream tasks, such as shape reconstruction\cite{paschalidou2019superquadrics,he2023dealing} and editing~\cite{tertikas2023generating,han2023improving}, object detection~\cite{liu2020dispersion,liu2019dispersion,li2023steering} and segmentation\cite{gao2022data,gao2023training,liu2022transfusion,liu2021refined,liu2021label,chang2022deeprecon,zhangli2022region,martin2023deep}.
Recent methods utilize deep neural networks to decompose objects into primitives \cite{paschalidou2019superquadrics,tulsiani2017learning,deng2020cvxnet}. \textcolor{black}{These primitive-based methods interpret a shape as a union of simple parts (\eg, cuboids, spheres, or superquadrics), offering explainable abstraction of the object shape.
To achieve high reconstruction accuracy, these methods require joint optimization of a large number of primitives which do not correspond often to the object parts and therefore limit the interpretability of the output. Therefore, devising methods that can discover a fewer number of primitives for efficiency, robustness and improved abstraction of complex shapes is an active research area. The use of fewer primitives to estimate complex object shapes with abstraction requires the discovery of primitives with broader and interpretable robust parametrization. }



In this paper, we investigate salient and efficient primitive descriptors to address flexible and explainable shape abstractions for complex objects with a minimal number of primitives. We take our inspiration from the physics-based deformable models (PDMs) \cite{metaxas2012physics}, which are capable of estimating and representing object shapes with strong abstraction ability and have been successfully applied to shape modeling in natural scenes, medical imaging, and graphics. A major issue of PDMs is that they rely on prior knowledge (\ie, handcrafted parametric initialization and optimization) for specific shape abstractions, which limits the usage of PDMs for general automated shape modeling. 
To address these limitations, we augment PDMs with strong abstraction ability and integrate them into a learning-based framework, named {\it Deep Deformable Models (DDMs)}, as illustrated in Fig.~\ref{cover_framework}. 
 
Compared to the traditional PDMs, we make use of deep neural networks to learn geometric representations of object shapes and overcome the parametric initialization limitation. 
\textcolor{black}{To enhance the shape coverage of DDM, we employ a diffeomorphic mapping that preserves shape topology to predict local non-rigid deformations for shape details beyond the coverage of global deformations. Integrated with the differentiable and invertible global deformations, our model is capable of learning semantic consistency for accurate abstractions.}
DDM also uses the PDM notion of  ``external forces'' to minimize the divergence between the predicted primitives and the target shapes during training~\cite{metaxas2012physics}. Note that the forces we reference in the paper aren't real forces we encounter in physical systems.
Instead, these are ``virtual forces'' computed based on the virtual displacement of the surface points of the primitive. This allows us to use kinematic formulations and Jacobians to regularize each sub-transformation during the primitive fitting.

To evaluate the proposed DDM, we conducted extensive experiments covering various problem settings on the shape abstraction task. We also show the improved abstraction accuracy, consistent semantic correspondence across the same shape category, and interpretable visualization results on \textit{ShapeNet}, compared to SOTAs. 

Our main contributions are summarized as follows:
\begin{itemize}
\item To the best of our knowledge, DDM is the first work that integrates physics-based deformable models with deep learning for accurate shape abstractions. 

\item We propose a generalized primitive formulation with differentiable global and local deformations to significantly improve the representation power of the primitive over the baselines. In addition, we propose to use kinematics-inspired losses to jointly regularize each sub-transformation of the primitive deformation.

\item Extensive experiments show that our method achieves better reconstruction accuracy and improved semantic correspondence compared to the state-of-the-art.
\end{itemize}

\begin{figure*} [t]
\begin{center}
\includegraphics[width=1\linewidth]{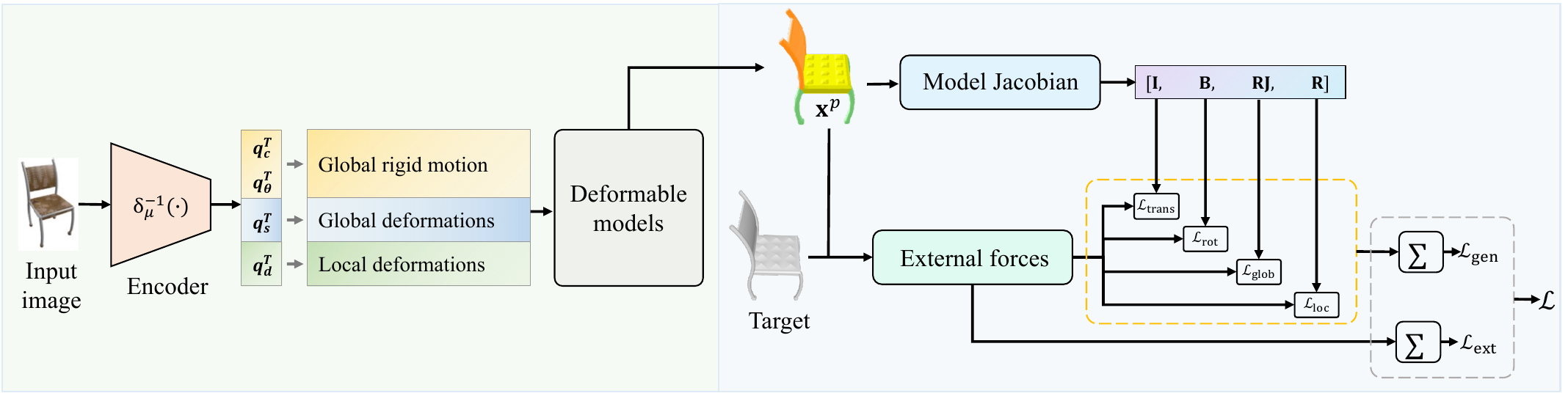}
\end{center}
   \caption{\textbf{Overview of training Deep Deformable Model (DDM).} Given an input image, DDM predicts a set of shape-related parameters $\textbf{q}_c$, $\textbf{q}_\theta$, $\textbf{q}_s$, and $\textbf{q}_d$ that describe global rigid motion, global and local deformations of the deformable model. For each reconstructed primitive of the deformable model ${{\textbf{x}}^p }$, DDM calculates the external forces and the model Jacobian matrices which transform the external forces from the data space to the latent parameter space. The generalized force loss ${\cal L}_\text{gen}$ optimizes the deformable model parameters in terms of translation ${{\cal L}_\text{trans}}$, rotation ${{\cal L}_\text{rot}}$, global deformations ${{\mathcal L}_\text{glob}}$ and local deformations ${{\cal L}_\text{loc}}$. The training loss is a weighted summation of both the external force loss ${\cal L}_\text{ext}$ and the generalized force loss ${\mathcal L}_\text{gen}$.}
\label{flowchart}
\end{figure*}

\section{Related Work}

\textcolor{black}{3D shape representation can be categorized into several mainstreams: (1) Voxel-based methods~\cite{choy20163d,wu2016learning} leverage voxels to capture 3D object geometry. These methods usually require large memory and computation resources. Some methods reduce the memory cost~\cite{maturana2015voxnet}, but the implementation complexity of these methods increases significantly. (2) Point cloud methods~\cite{fan2017point,qi2017pointnet++} require less computation, but additional post-processing is necessary to address the lack of surface connectivity for mesh generation. (3) Mesh-based methods~\cite{liao2018deep,groueix2018papier} can yield smooth shape surfaces, but most of them lack output interpretability (abstraction ability). (4) Implicit function-based methods~\cite{mescheder2019occupancy,chen2019learning,park2019deepsdf} can also reconstruct shapes with high accuracy, but
they require heavy post-processing to extract meshes. (5) Primitive-based methods~\cite{tulsiani2017learning,paschalidou2019superquadrics} represent object shapes by deforming a number of primitives, each of which is explicitly defined by a set of shape-related parameters.}

\textcolor{black}{\textbf{Unsupervised primitive-based methods.}
Our approach falls into unsupervised primitive-based shape abstractions which have been revisited in deep learning and have recently demonstrated promising results. Paschalidou~\etal~\cite{paschalidou2019superquadrics} developed a method that combines superquadrics with deep networks. Given the number of primitives used, it estimates a set of superquadric parts that enable 3D shape parsing. This method has been further extended to estimate hierarchical parts from 3D data \cite{paschalidou2020learning}. Other shapes such as cuboids \cite{tulsiani2017learning,niu2018im2struct,zou20173d}, spheres \cite{hao2020dualsdf,paschalidou2021neural} and convexes \cite{deng2020cvxnet} have also been used for primitive-based reconstruction. However, these basic parts only offer limited shape coverage and cannot address the accurate estimation of complex shapes with semantically meaningful part decomposition. Compared to these methods, DDM is capable of abstracting complex object shapes using a small number of primitive with improved geometry coverage. }

\textbf{Implicit function-based methods.} This set of methods mainly leverage implicit functions (\ie, level-sets) to directly estimate the signed distance function~\cite{mescheder2019occupancy,chen2019learning,park2019deepsdf,liu2021label}. While they achieve high reconstruction accuracy, they require heavy post-processing (\eg, marching cubes) to recover the shape surface. In contrast, primitive-based methods seek to decompose a target shape into semantic parts and also decompose each part into explicit shape-related parameters (\eg, scaling, squareness, tapering, bending), which contribute to the understanding of primitive deformation. These shape-related parameters enable explicit modeling for each shape part without any post-processing and provide semantic consistency among shapes.
 
\textcolor{black}{\textbf{3D correspondence learning.} We note that some implicit function-based approaches~\cite{deng2021deformed,zheng2021deep,halimi2019unsupervised,liu2020learning,cheng2021learning} work on the task of learning 3D correspondence, aiming at finding corresponding points in the target instance for points of the source instance. They require pair-wise 3D source and target instances for training,
which differs from our motivation.
Our method learns parameters of primitive deformations in order to abstract the 3D shape from the input image. Once trained with individual samples of the same shape category, our model can abstract each shape instance and naturally build correspondence between such instances as a byproduct, which does not require any paired data as input.}

\textbf{Parameterized deformable models.}
Prior research works developed parameterized deformable models that abstracted multiple shapes with relatively few parameters. A notable example is the work of \cite{kass1988snakes} which exploited computational physics in the modeling process and proposed snakes, a locally parameterized deformable model. The snake formulation employs
a force field computed from data to fit the model. Nevertheless, snakes which use locally defined deformations do not intrinsically offer shape abstractions.
The problem of shape abstraction was partially addressed by using superquadric ellipsoids that can deform using a few global parameters~\cite{pentland1987perceptual},
allowing a simpler deformable model with fewer parameters to represent the target object. They further apply polynomial approximation to their model and develop an efficient and useful deformation model to reconstruct smooth and symmetrically deformed parts of the object. However, as complex shapes are required to reconstruct or segment, this approach is less efficient than the finite element solution with one node~\cite{zienkiewicz1977finite}.
To overcome the limitations of locally deforming models that do not offer shape abstraction, and globally deforming models that have limited shape representing,
\cite{terzopoulos1991dynamic} developed a new physics-based framework offering multiscale global and local deformations, and demonstrated its power using deformable superquadrics. 
Although their framework was able to address complex shapes and motion estimations of objects, it relies on handcrafted parameter initialization~\cite{jones1998image}.

\section{Method}
Given an input image to be reconstructed, the goal of our method is to incorporate a differentiable deformable model to predict $P$ primitives that best describe the target shape. Each primitive is explicitly represented by a set of shape-related parameters ${\textbf{q}}$ with global and local deformations.
The overview of our model training is given in Fig.~\ref{flowchart}.

\subsection{Geometry and primitive parameterization} 
\label{geometry}
\textcolor{black}{We begin by summarizing the concept of PDMs and, along the way, introduce the notations. Geometrically, DDM models each deformable primitive as a closed surface with a model-centered coordinate $\phi$. As shown in the deformation geometry of Fig.~\ref{cover_framework}, given a point on the primitive surface, its location $\textbf{x} = (x,y,z)$ \textit{w.r.t.} the world coordinate $\Phi $ is:
\begin{equation}
    \textbf{x} = {\textbf{c}} + {\textbf{R}} \textbf{p} = {\textbf{c}} + {\textbf{R}} (\textbf{d} + \textbf{s}),
\label{eq1}
\end{equation}
where ${\textbf{c}}$ and $\textbf{R}$ represent the translation and rotation \textit{w.r.t.}~the world coordinate $\Phi$, respectively;  $\textbf{p}$ denotes the relative position of the point on the primitive surface \textit{w.r.t.}~$\phi$, which includes global deformation $\textbf{s}$ and local deformation $\textbf{d}$. Global deformations are expected to efficiently capture salient features of the target shape using a minimum number of parameters, while local deformations allow the model to represent the fine-scale details.} We denote the learnable parameters for the translation and rotation as $({\textbf{q}}_c, {\textbf{q}}_\theta)$, where ${\textbf{q}}_c = \textbf{c}$ and ${\textbf{q}}_{\theta }$ is a four-dimensional quaternion related to $\textbf{R}$ defined in~\cite{terzopoulos1991dynamic}.

\textbf{Primitive formulation.}
We employ superquadrics as our basic primitive formulation for the global deformation $\textbf{s}$. Each superquadric surface ${\textbf{e}}$ is explicitly defined by a set of shape-related parameters:
\begin{equation}
{\textbf{e}} = {a_0} \left[ {\begin{array}{*{20}{l}}
{{a_1}\cos^{{\varepsilon _1}} u \cos^{{\varepsilon _2}}v}\\
{{a_2}\cos^{{\varepsilon _1}} u \sin^{{\varepsilon _2}}v}\\
{{a_3}\sin^{{\varepsilon _1}}u}
\end{array}} \right],
\end{equation}
where $-\pi/2 \leq u \leq \pi/2,
-\pi \leq v \leq \pi$.
Here, ${a_0}$ is a scaling parameter; ${a_1}$, ${a_2}$, ${a_3}$ denote the aspect ratio for $x$-, $y$-, $z$- axes, respectively; and ${\varepsilon _1},{\varepsilon _2}$ are squareness parameters. 

\textbf{Global deformations.}
To improve the geometric coverage of these primitives, we introduce parameterized tapering and bending deformations. These additional global deformations are defined as continuously differentiable and commutative functions following \cite{metaxas2012physics}. Specifically, due to their suitability for natural objects, we integrate linear tapering and bending of the superquadric ${\textbf{e}} = {({e_1},{e_2},{e_3}) ^\top}$ into one parameterized deformation $\textbf{T}$ and give the formulation of the reference shape as:
\begin{align}
    {{\textbf{s}}} &= {\textbf{T}}({\textbf{e}},{t_1},{t_2},{b_1},{b_2},{b_3}) \\
          &= \left( {\begin{array}{*{20}{l}}
{(\frac{{{t_1}{e_3}}}{{{a_0}{a_3}}} + 1){e_1} + {b_1}\cos (\frac{{{e_3} + {b_2}}}{{{a_0}{a_3}}})\pi {b_3}}\\
{(\frac{{{t_2}{e_3}}}{{{a_0}{a_3}}} + 1){e_2}}\\
{{e_3}}
\end{array}} \right),
\end{align}
where ${t_1}$ and ${t_2}$ are the tapering parameters; ${b_1}$, ${b_2}$, and ${b_3}$ are the magnitude, location, and influence region of bending, respectively. The learnable parameters for $\textbf{s}$ are then denoted as ${\textbf{q}}_s = (a, \epsilon, t, b)$, where $a = (a_0, a_1, a_2, a_3), \varepsilon = (\varepsilon _1, \varepsilon _2)$, $t = (t_1, t_2)$, and $b = (b_1, b_2, b_3)$.

\textcolor{black}{\textbf{Diffeomorphic local deformations.} }
We use local deformations to capture fine details beyond the coverage of global deformations. Previous approaches~\cite{metaxas1992physics} adopted the finite element method~\cite{zienkiewicz1977finite} to estimate local deformations. This requires the handcrafted design of shape functions for the chosen fine elements with additional computational costs for accurate local deformation estimation. \textcolor{black}{In this paper, we introduce a diffeomorphic mapping to estimate the local deformations $\textbf{d}$. Due to the differentiable and invertible properties of diffeomorphism, it preserves topology and guarantees one-to-one mapping during deformations~\cite{dalca2018unsupervised}. 
In addition, since the global deformations used are invertible, the composed deformation of global and local deformations in our model is invertible and smooth,  which thus facilitates the learning of semantic correspondences for shape abstraction. To be specific, given the encoded local feature $l$ from the encoder $\delta _\mu ^{ - 1}({\cdot})$, we first use a convolution layer to map $l$ to a vector field $v_0$, and then map $v_0$ to a stationary velocity field (SVF) $v$ using a Gaussian smoothing layer. $v$ is defined via the ordinary differential equation~\cite{arsigny2006log}: 
\begin{equation}
\frac{{d{\psi ^{(\tau)}}}}{{d\tau}} = v({\psi ^{(\tau)}}),
\label{svf}
\end{equation}
where $\psi ^{(\tau)}$ is the path of diffeomorphic non-rigid deformation field parameterized by $\tau \in [0,1]$ and $\psi ^{(0)}= \textbf{I}$ is an identity transformation. To obtain the final local non-rigid deformation $\textbf{d} = \psi^{(1)}$ at time $\tau=1$, we follow~\cite{arsigny2006log,dalca2018unsupervised} and employ an Euler integration with a scaling and squaring layer (SS) to solve Eq.~(\ref{svf}) and predict the learnable local deformation parameters $\textbf{q}_d = \textbf{d}$}.

\textbf{Kinematics and dynamics.}
In our modeling paradigm, the deformable primitive continuously deforms from an initial shape (\eg, a sphere) to the target shape using the Lagrangian equations of motion given as:
\begin{equation}
    \textbf{M} \ddot {\textbf{q}} + \textbf{C} \dot {\textbf{q}} + \textbf{K} \textbf{q} = g_q + f_q,
\end{equation}
where $\cdot \cdot $ denotes the second-order time derivative; $\textbf{M}, \textbf{C}, \textbf{K}$ are the mass, damping, and stiffness matrices, respectively; $g_q$ is the inertial forces generated from the dynamic coupling between the local and global deformations; $f_q$ is the generalized forces which we will explain next. In this paper, we set $\textbf{M}=\textbf{0}$, $\textbf{C}=\textbf{1}$, $\textbf{K}=\textbf{0}$, $g_q=0$ and use a simplified Lagrangian dynamic model given as:
\begin{equation}
    \dot {\textbf{q}}  = f_q.
\end{equation}

From Eq. (\ref{eq1}), we can derive the velocity of a point on the primitive surface as:
\begin{equation}
{\dot {\textbf{x}}} = \dot{\textbf{c}} + \dot {\textbf{R}} {{\textbf{p}}} + {\textbf{R}} \dot {\textbf{p}}= \dot {\textbf{c}} + {\textbf{B}} \dot \theta  + \textbf{R} \dot{\textbf{s}} +{\textbf{RS}}\dot{\textbf{q}}_d,
\label{x_dot}
\end{equation}
where $\cdot $ denotes the first-order time derivative; ${\textbf{B}} = {{\partial {\textbf{Rp}}}}/{\partial {\theta }}$, with $\theta = \textbf{q}_\theta$ the rotational coordinates and $\dot {\textbf{s}} = [\partial {\textbf{s}} / \partial \textbf{q}_s]{\dot {\textbf{q}}}_s = \textbf{J} \dot{\textbf{q}}_s$, with \textbf{J} the Jacobian matrix of the model-centered coordinates $\phi$ \textit{w.r.t.}~the global deformation parameters at each point. We set the shape matrix \textbf{S} to an identity matrix \textbf{I} in DDM since we use the one-to-one mapping for local deformation estimation. We note that the size of the Jacobian matrix is determined by the type of global deformations used.
Eq.~(\ref{x_dot}) can be further written in the form:
\begin{equation}
    \dot {\textbf{x}} = [{{\textbf{I}}, \textbf{B} , \textbf{R}  \textbf{J} , \textbf{R} }] \dot{\textbf{q}}= {\textbf L}  \dot{\textbf{q}},
\label{x_dot_l}
\end{equation}
where $\textbf{L}$ is the deformable model's Jacobian matrix that includes the Jacobians ${\textbf{J}}$ for translation, rotation, and deformations \cite{metaxas2012physics}.

Eq.~(\ref{x_dot_l}) shows how a change in a 3D point $\textbf{x}$ is translated to a change in the shape-related parameters of the primitive. Our approach is inspired by the kinematics of PDMs, which takes the shape deformation as a dynamic system and thus the time derivative is used for $\textbf{x}$ and $\textbf{q}$. In our setting, at an arbitrary training iteration $t$, the 3D point $\textbf{x}$ from the primitive surface should overlap with the point from the surface of the ground-truth shape. $ \dot {\textbf{x}}$ (\textit{a.k.a} $d \textbf{x}$) shows how this point should change to make this possible and $\dot{\textbf{q}}$ (\textit{a.k.a} $d \textbf{q}$) is the corresponding change in the shape-related parameters of the primitive. 
In the following, we will show that using Eq.~(\ref{x_dot_l}), we can convert the minimization of 3D point-wise difference (external forces) between the primitive and the target shape to the minimization of the parameter-wise difference (generalized forces), making it possible to explicitly supervise the learning of each sub-transformation component, \ie, translation, rotation, global deformation, and local deformation.



\subsection{DDM training and network losses} \label{optimization}
In our training strategy, in addition to directly sampling points from the primitive surface and optimizing them,  we seek to optimize groups of shape-related parameters that control the transformations of the primitive, \ie, translation, rotation, global and local deformations. 
Therefore, we define the following loss function to train and optimize DDM:
\begin{equation}
{\cal L} = {\lambda_\text{ext}{\cal L}_\text{ext}} + {\lambda_\text{gen}{\cal L}_\text{gen}},
\label{loss}
\end{equation}
which is a weighted summation of the loss ${\cal L}_\text{ext}$ computed using external forces from the \textbf{data space} and the loss ${\cal L}_\text{gen}$ computed using generalized forces from the \textbf{latent parameter space}; $\lambda_\text{ext}$ and $\lambda_\text{gen}$ are their weights, respectively. 

\textbf{External model loss.}
To fit the primitives to the target shape, we train an encoder $\delta _\mu ^{ - 1}({\cdot})$ to optimize the loss ${{\mathcal L}_{{\text{ext}}}} $ computed using the external forces applied to the primitives:\textcolor{black}{
\begin{equation}
{{\mathcal L}_{{\text{ext}}}} = \sum\limits_{p} {\sum\limits_r {f_r^p} }  = \gamma \sum\limits_{p = 1}^{ P} {\sum\limits_{{r} \in {{\cal  M}_p}}  {\cal D}({{{\cal  M}_p,{\cal T}} }) }. 
\end{equation}
${\cal D}({{{\cal  M}_p,{\cal T}} }) $ is the distance of the points on the target shape ${\cal T}$ to the points $r$ on the $p$-th predicted primitive ${\cal  M}_p$, where $P$ is the total number of used primitives and $\gamma$ is the strength factor for the external forces $f_r^p$.} The external force loss ${{\mathcal L}_{{\text{ext}}}} $ measures how well the primitives are deformed to fit the target shape in the data space during training.

\begin{figure}[t]
  \begin{center}
    \includegraphics[width=1\linewidth]{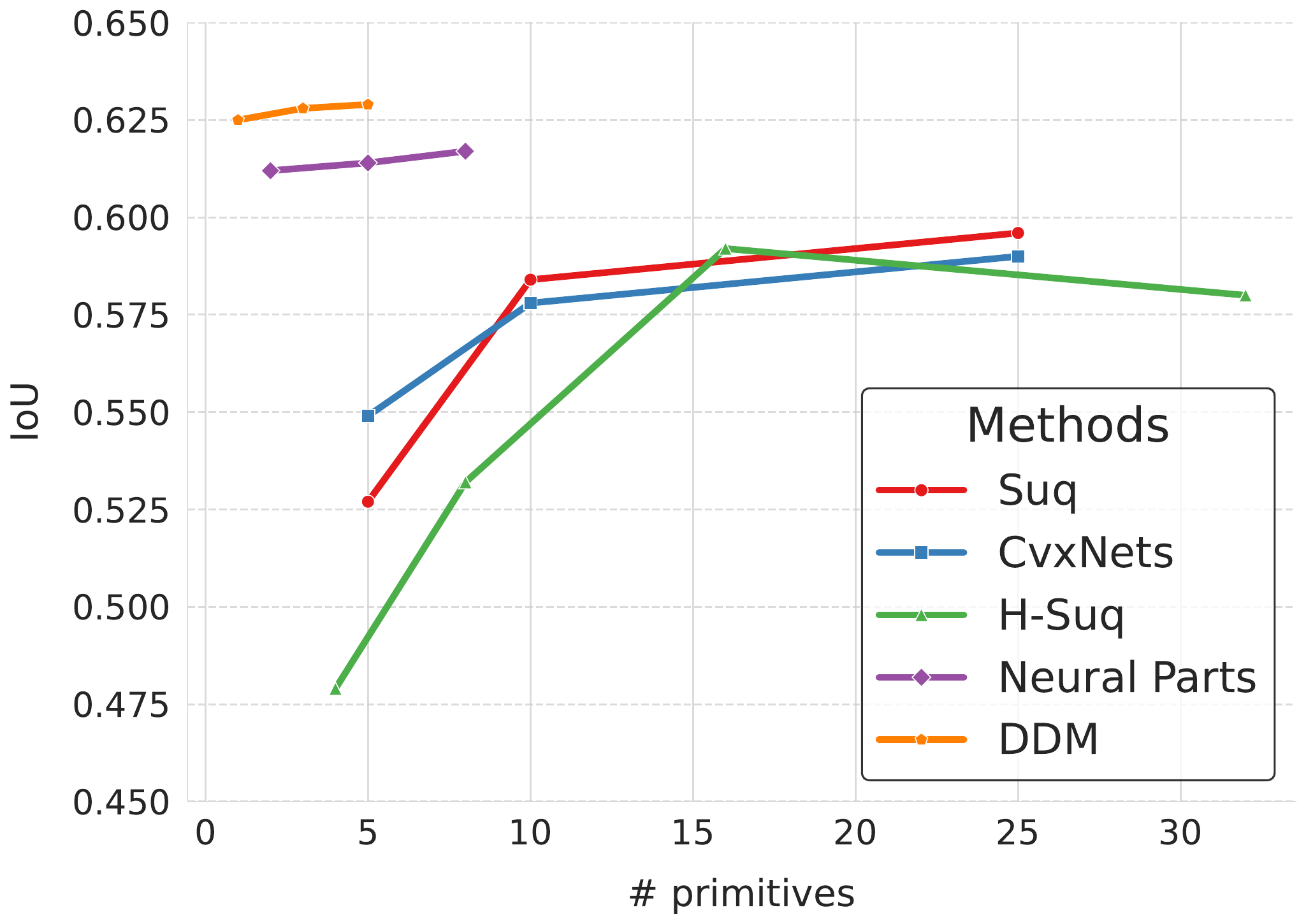}
  \end{center}
  \caption{\textbf{Analysis of the reconstruction accuracy \textit{v.s.}~the number of primitives used.} We compare our reconstruction accuracy with those of other primitive-based methods on \textit{ShapeNet}, and observe better performance with only a small number of primitives.}
\label{multi_prim}
\end{figure}

\begin{figure*} [t]
\begin{center}
\includegraphics[width=1\linewidth]{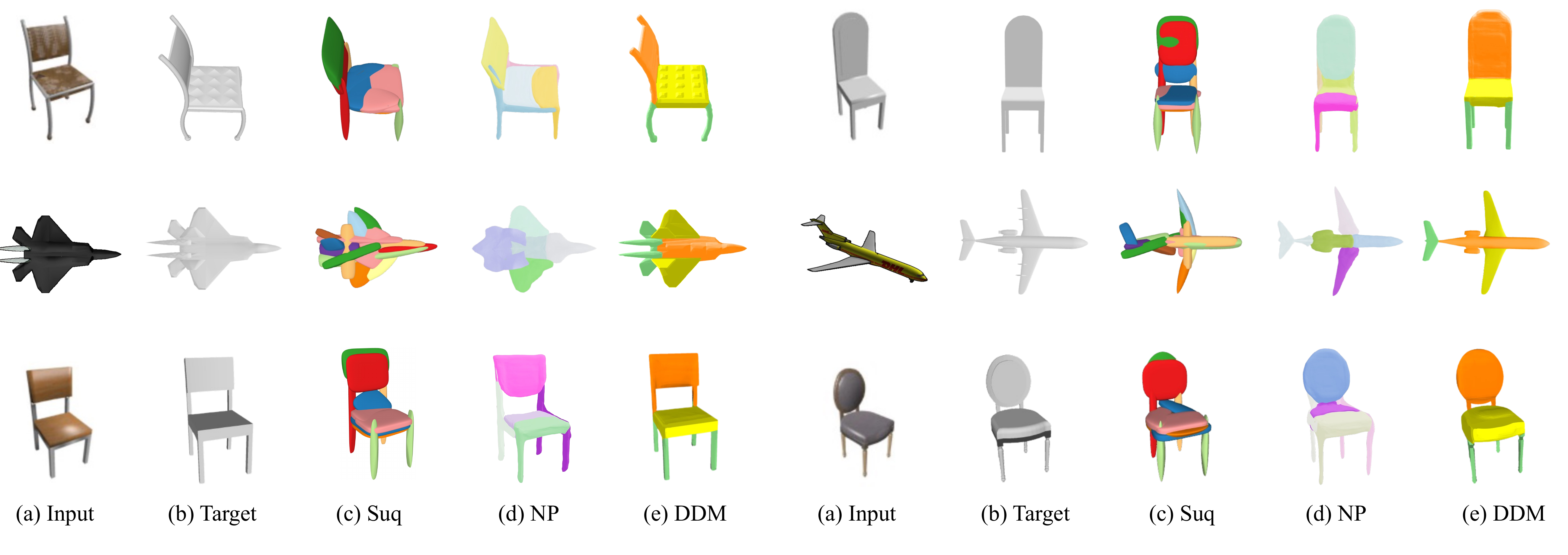}
\end{center}
\vspace{-6pt}
\caption{\textbf{Visual results on \textit{ShapeNet}.} (a) Input image, (b) Target meshes, (c) Suq \cite{paschalidou2019superquadrics} with around 20 primitives, (d) Neural Parts \cite{paschalidou2021neural} with 5 primitives, and (e) DDM using 3 and 6 primitives for chairs and airplanes, respectively. We observe DDM outperforms the other methods in terms of reconstruction accuracy, and achieves significantly better part consistency.}
\label{cover_img}
\end{figure*}

\textbf{Generalized model loss.} 
Given Eq.~(\ref{x_dot_l}), using the principle of \textcolor{black}{virtual work}~\footnote{\textcolor{black}{In mechanics, virtual work is the total work done by the applied forces on a mechanical system as it moves through a set of virtual displacements.}}, we can determine the relationship between the generalized forces and the external forces. In particular, the energy of the $p$-th primitive, ${{\cal E}^p_f}$, \textit{i.e.}, the amount of virtual work required to deform a primitive so that it aligns with a target shape, is expressed as:
\begin{equation}
{{\cal E}^p_f}  = \int {{(f^p) ^\top}d{\textbf{x}}^p = \int {{(f^p) ^\top}\textcolor{black}{{\textbf{L}}^p}d{\textbf{q}}^p} = \int {{f^p_q}d{\textbf{q}}^p}},
\label{L}
\end{equation}
where $f^p_q$ is the generalized forces applied to ${\cal  M}_p$ and is computed using the external forces $f^p$ and the model Jacobian matrix ${\textbf{L}}^p$. 
This allows us to employ the generalized forces $f^p_q$ in the latent parameter space to facilitate the primitive prediction. Specifically, given the model Jacobian ${\textbf L}^p = [{{\textbf{I}}^p, \textbf{B} ^p, \textbf{R} ^p \textbf{J} ^p, \textbf{R} ^p}]$~\cite{metaxas2012physics}, we express $f^p_q$ as:
\begin{equation}
\begin{split}
{f_q^p} = {{({f^p}) ^\top}{{\textbf{L}}^p}} & = [{{{(f^p)} ^\top}}, {{{(f^p)} ^\top}{\textbf{B}}^p}, {{{(f^p)} ^\top}{\textbf{R}}^p{{\textbf{J}}^p}}, {{{(f^p)} ^\top}{\textbf{R}}^p}] \\
 & = {[(f^p_{c}) ^\top,(f^p_{\theta})  ^\top,(f^p_{s})  ^\top,(f^p_{d}) ^\top} ], 
\end{split}
\end{equation}
where $f^p_{c}$ and $f^p_{\theta}$ represent the generalized forces for the translation and rotation; $f^p_{s}$ and $f^p_{d}$ represent the generalized forces for the global and local deformations. In our learning framework, in addition to the external forces, we also train the encoder $\delta _\mu ^{ - 1}({\cdot})$ to optimize these four generalized force components and define the generalized model loss ${\cal L}_\text{gen}$ as:

\begin{equation}
{{\cal L}_\text{gen}} = {{\cal L}_\text{trans}} + {{\cal L}_\text{rot}} + {{\cal L}_\text{glob}} + {{\cal L}_\text{loc}},
\end{equation}
where 
\begin{align*}
{{\mathcal L}_\text{trans}} &= \sum\limits_{p = 1}^{P} {(f_{c}^p} {) ^\top} = \sum \limits_{p = 1}^{P} {\sum\limits_r {{{(f_r^p)} ^\top}} } ,    \\  {{\cal L}_\text{glob}} &= \sum\limits_{p = 1}^{P} {(f_{s}^p} {) ^\top} = \sum\limits_{p = 1}^{P} {\sum\limits_r {{{(f_r^p)} ^\top}{\textbf{R}}^p{{\textbf{J}}^p}} } ,       \\     
{{\cal L}_\text{rot}} &=  \sum\limits_{p = 1}^{P} {(f_{\theta} ^p} {) ^\top} = \sum\limits_{p = 1}^{P} {\sum\limits_r {{{(f_r^p)} ^\top}{\textbf{B}}^p} } ,        \\  {{\cal L}_\text{loc}} &= \sum\limits_{p = 1}^{P} {(f_{d}^p} {) ^\top} = \sum\limits_{p = 1}^{P} {\sum\limits_r {{{(f_r^p)} ^\top}{\textbf{R}}^p} } , 
\end{align*}
are the generalized model losses associated with the translation, rotation, global, and local model degrees of freedom, respectively.
\textcolor{black}{Note that, by decomposing the generalized forces into different components and by minimizing each force component, we can directly optimize the corresponding deformation components, which provides a more strict regularization of the primitive fitting.}   

\section{Experiments}
\label{exp_3d}

\begin{figure*}[t]
  \begin{center}
    \includegraphics[width=\linewidth]{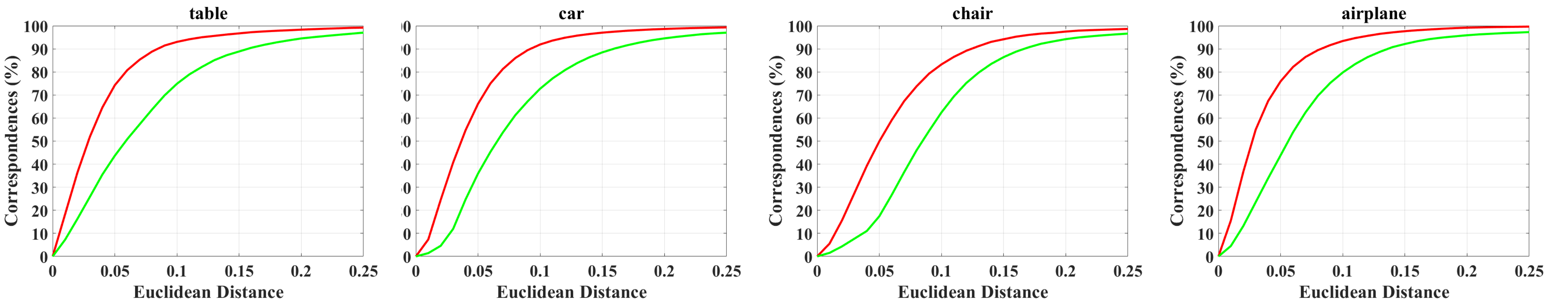}
  \end{center}
  \caption{\textbf{Correspondence results.} We evaluate DDM on both registered (red) and unregistered (green) data. We compute the distances from the transferred points to ground truth points and report the percentage of testing pairs where the distances are below a given threshold.}
\label{corr_chart}
\end{figure*}

\begin{figure*}[t]
  \begin{center}
\includegraphics[width=1\linewidth]{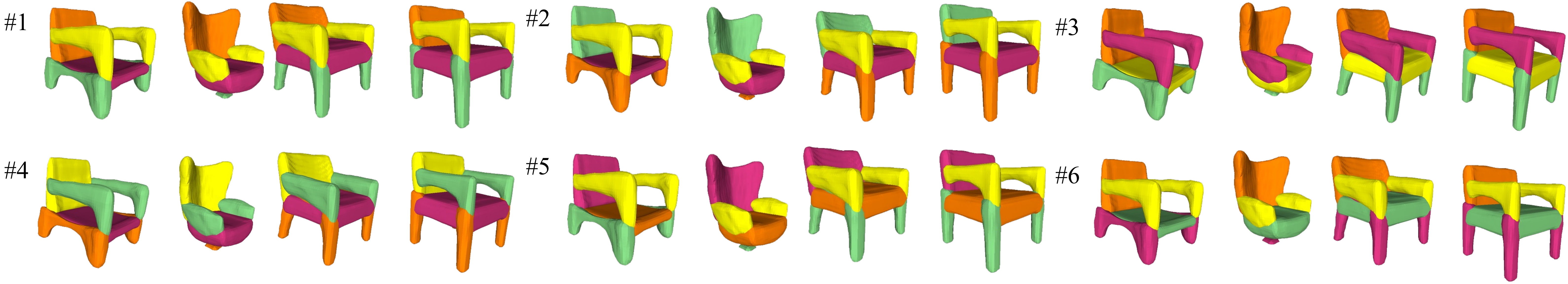}
  \end{center}
  \caption{\textbf{Sensitivity to initialization.} We train DDM with six different random seeds (indicated with \#) on the chair category of \textit{ShapeNet}. For each seed, we observe the consistent part correspondence (\eg, chair back, legs, and seat denoted as the same color), across the four ``chair'' shape instances.}
\label{ab_ini}
\end{figure*}

\begin{figure}[!htbp]
  \begin{center}
\begin{minipage}{1\linewidth}
  \begin{center}
    \includegraphics[width=\linewidth]{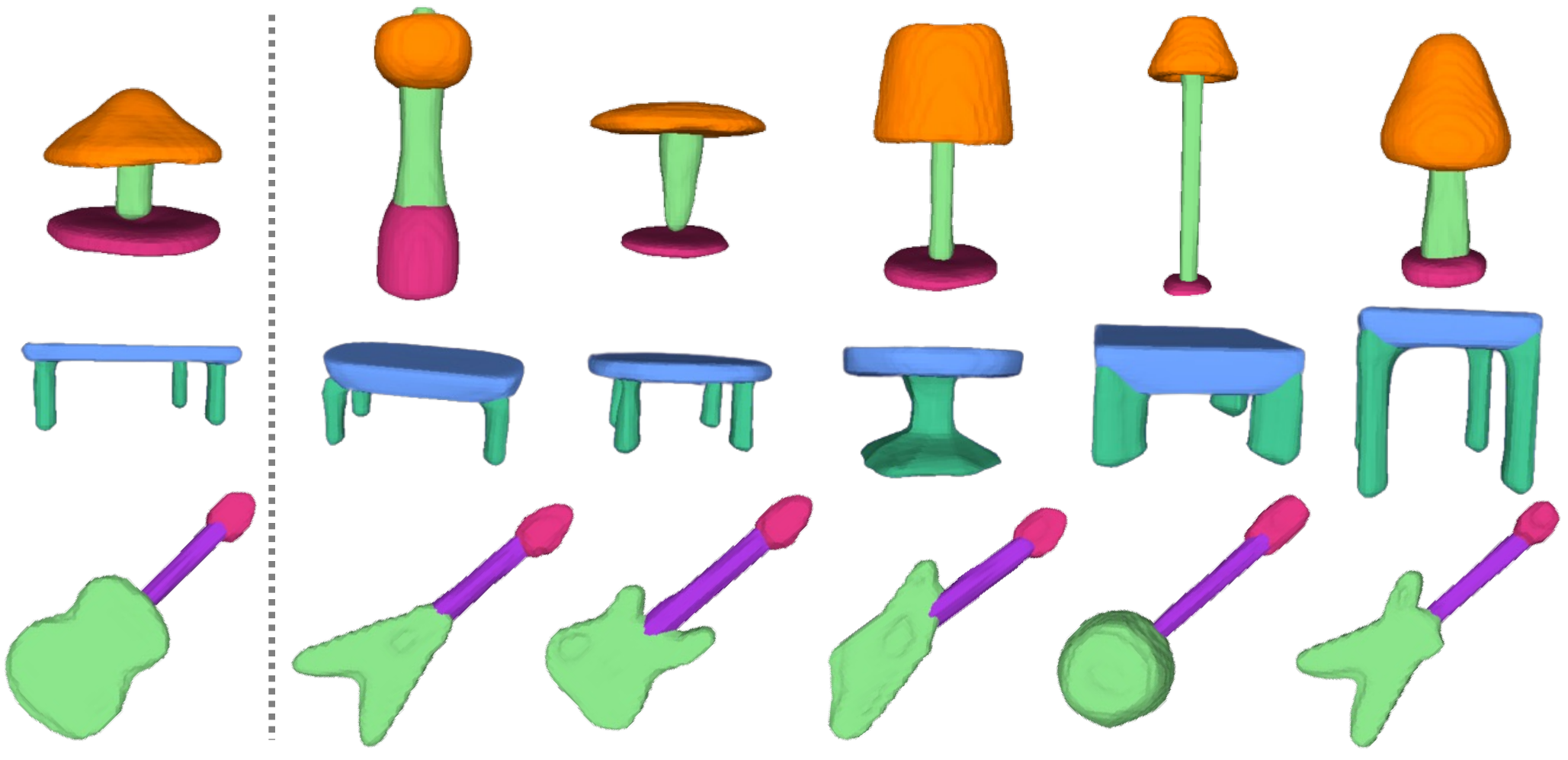}
  \end{center}
	\end{minipage}\\


  \end{center}
\caption{\textbf{Part label transfer results.} The first column is the source shape, while the rest is the transferred labels from source shapes through learned correspondences using DDM.
}
\label{part_transfer}
\end{figure}

\begin{table}[t]
\centering
\caption{\textbf{Quantitative results on \textit{ShapeNet}.} We evaluate DDM (4) against Suq ($\leq 64$)~\cite{paschalidou2019superquadrics}, CvxNets (25)~\cite{deng2020cvxnet}, H-Suq ($\leq 64$)~\cite{paschalidou2020learning}, and NP (5)~\cite{paschalidou2021neural}. We report IoU and Chamfer-$L_1$ distance for comparison. Numbers in ($\cdot$) indicate the number of primitives used.}
\label{Table3d}
\renewcommand\tabcolsep{12pt}
\resizebox{1\linewidth}{!}{
\begin{tabular}{l|ccccc}
\toprule
Method & airplanes & table & car & chair \\ 
\midrule
Suq & 0.456 & 0.180 & 0.650 & 0.176\\
CvxNets & 0.598 & 0.473 & 0.675 & 0.491\\
H-Suq & 0.529 & 0.491 & 0.702 & 0.526\\
NP & 0.611 & 0.531 & 0.719 & 0.532\\
\midrule
DDM & \textbf{0.631} & \textbf{0.544}  & \textbf{0.721} & \textbf{0.542}\\ 
\bottomrule
\end{tabular}} 
\end{table}
\subsection{Settings and datasets}
We evaluate the performance of DDM on \textit{ShapeNet}~\cite{chang2015shapenet}, a richly-annotated, large-scale dataset of 3D shapes. 
A subset of \textit{ShapeNet} including 50K models and 13 major categories are used in our experiments. 
We split the dataset into training and testing sets following~\cite{choy20163d}.
In all experiments, Adam \cite{kingma2014adam} is employed for optimization and the learning rate is initialized as $10^{-4}$. We use a batch size of 32 and train the model for 300 epochs. 
All experiments are implemented with PyTorch and run on a Linux system with eight Nvidia A100 GPUs. We draw 2K random samples from the surface of the target mesh, and sample 1K points for each generated primitive during training. During the evaluation, we uniformly sample 100K points on the target/predicted meshes for the calculation of the volumetric Intersection over Union (IoU) and Chamfer-$L1$ distance (CD). We use the standard ResNet-18~\cite{he2016deep} as the encoder ${\delta^{-1}_\mu }(\cdot)$ for all experiments.
The encoder output is followed by an average pooling and two fully connected layers to estimate four individual vectorized parameters that represent translation, rotation, global, and local deformations.

\subsection{Representation power}
We compare DDM with various primitive-based explicit representation baselines~\cite{paschalidou2019superquadrics,paschalidou2020learning,paschalidou2021neural,deng2020cvxnet} on \textit{ShapeNet}. In Fig.~\ref{multi_prim}, we report the reconstruction accuracy by varying the number of primitives and test the performance in terms of IoU. Our model shows leading reconstruction performance regardless of the number of primitives used. We also observe that the curve saturates when adding more primitives to our model. We attribute this to the broad geometric coverage of our primitive formulation.

\subsection{Reconstruction accuracy}
In this experiment, we train DDM with 4 primitives and train the baseline models following their reported experimental setups. Specifically, for Suq and H-Suq, we use a maximum number of 64 primitives; for CvxNets and Neural Parts, we report the results using 50 and 5 primitives, respectively, which in their papers lead to the best performance.
The quantitative results measured by IoU and Chamfer-$L_1$ distance are reported in Table \ref{Table3d}.
In \cref{cover_img}, we highlight the superiority of our approach in capturing complex geometry (\eg, high curvatures) of various object shapes.
We find that while some baseline methods use multiple primitives with obvious overlap to abstract the object shapes, DDM can capture the complete geometry of the chairs and airplanes with better details using a small number of primitives. Moreover, DDM demonstrates meaningful semantic correspondence among individual instances from the same category (see the parts of the four chairs with the same color codes), indicating clear advantageous interpretability.

\subsection{Keypoint and part transfer}
To quantitatively evaluate the semantic correspondence, we carry out a keypoint transfer task on \textit{KeypointNet}~\cite{you2020keypointnet} due to the lack of ground truth correspondence on \textit{ShapeNet}. We follow the settings of \cite{liu2020learning,chen2020unsupervised,cheng2021learning} and evaluate on both registered and unregistered data. We compute the distances from the transferred points to ground truth points, and report the percentage of testing pairs where the distances are below a given threshold in Fig.~\ref{corr_chart}. 
We also validate our approach on the part label transfer task \cite{wang2020few}. Following \cite{deng2021deformed}, we use five labeled shapes from \textit{ShapeNet-Part} \cite{yi2016scalable} dataset as source shapes, and transfer their labels to other instances from the same category via the learned correspondences. The results in Fig.~\ref{part_transfer} illustrate the semantic consistency across instances from the same category, \eg, the lampshades are always matched despite large variances of the object structures.

\subsection{Ablation study}
\label{ab}
We first investigate the effect of loss components and parameterized deformations (global and local) in terms of reconstruction accuracy. In Table~\ref{ab_table}, using the ``leave-one-out'' way, each of the loss terms and the deformations are highlighted and demonstrated to be a uniquely effective component within our full model. 
We also investigate the sensitivity of DDM to primitive initialization. We demonstrate our approach generates identical semantic partitions by using different random initialization. Specifically, we train our model with six different random seeds, and observe in Fig.~\ref{ab_ini} that the predictions preserve similar shape parts, and thus semantic meaningful.

\begin{table}[!htbp]
\centering
\caption{\textbf{Ablation study on losses and deformations.} We report Chamfer-$L_1$ Distance (CD) and IoU on \textit{ShapeNet}.}
\renewcommand\tabcolsep{10pt}
\renewcommand{\arraystretch}{1}
\resizebox{0.47\textwidth}{!}{
\begin{tabular}{cccc|cc}
\toprule
\cmidrule(lr){1-6}
   ${{\cal L}_\text{ext}}$&  ${{\cal L}_\text{gen}}$ & global & local  &CD ($\downarrow$) & IoU ($\uparrow$) \\ 
\midrule
 \xmark & \cmark  & \cmark  & \cmark  & 0.181 &  0.531 \\
\cmark  &  \xmark & \cmark  & \cmark & 0.125 &  0.562 \\
\midrule
\cmark & \cmark  & \xmark & \cmark  & 0.136 &  0.597 \\
\cmark  & \cmark  & \cmark  & \xmark  & 0.109 &  0.621 \\
\midrule
\cmark  & \cmark  & \cmark  & \cmark  & \textbf{0.097} &  \textbf{0.629} \\
\bottomrule
\end{tabular}} 
\label{ab_table}
\end{table}

\section{Conclusion}
In this work, we introduced a novel kinematics-inspired learning approach for improved object shape abstractions. The generalized primitive formulation allows the proposed model to accurately capture the geometric structures of object shapes using a small number of shape components. Moreover, our kinematics-inspired modeling provides multiscale parameterized shape representation ability while preserving the semantic interpretation of the shape. Extensive experiments demonstrate that our approach yields both accurate and explainable shape abstractions across various tasks. Our future work will consider including more primitive definitions (\eg, multigenous primitives) and global deformations (\eg, shearing, twisting) to enhance the expressiveness of our primitives in more general and complex shape abstraction scenarios.

\bibliography{aaai24}

\end{document}